\documentclass[twocolumn,10pt]{article}

\usepackage{latex8,times,latexsym,epsfig,fancyheadings,alltt,a4wide}

\usepackage{url}
\def\sburl#1{[\url{#1}]}

\def\expr#1{\texttt{#1}}
\def\predicate#1{\texttt{#1}}

\newcommand{\queryseq}{-$>$\ }
\newcommand{\querydom}{\^{}\ }
\newcommand{\queryassoc}{$=>$\ }

\def\smtt#1{{\small\tt #1}}

\newenvironment{sv}{\small\begin{alltt}}{\end{alltt}\normalsize}

\def\arXivhack{\vspace{-6pt}}

\title{Querying Databases of Annotated Speech}
\author{Steve Cassidy\\
Department of Linguistics\\
Macquarie University\\
Sydney, NSW 2109,\\
Australia\\
\smtt{steve.cassidy@mq.edu.au}\\
\and 
Steven Bird\\
Linguistic Data Consortium,\\
University of Pennsylvania, \\
3615 Market St, Suite 200,  \\
Philadelphia, PA 19104-2608, USA \\
\smtt{steven.bird@ldc.upenn.edu}
}

\begin{document}

\maketitle
\thispagestyle{empty}

\begin{abstract}
  Annotated speech corpora are databases consisting of signal data
  along with time-aligned symbolic `transcriptions'.  Such databases
  are typically multidimensional, heterogeneous and dynamic.  These
  properties present a number of tough challenges for representation
  and query.  The temporal nature of the data adds an additional layer
  of complexity.  This paper presents and harmonises two independent
  efforts to model annotated speech databases, one at Macquarie
  University and one at the University of Pennsylvania.
  Various query languages are described, along
  with illustrative applications to a variety of analytical problems.
  The research reported here forms a part of several ongoing projects
  to develop platform-independent open-source tools for creating,
  browsing, searching, querying and transforming linguistic databases,
  and to disseminate large linguistic databases over the internet.
\end{abstract}

\Section{Databases of Annotated Speech Recordings}

Annotated corpora have been an essential component of research and
development in language-related technologies for some years.
Text corpora have been used for developing information retrieval
and summarisation software (e.g. MUC \cite{MUC7}, TREC \cite{Voorhees98}),
automatic taggers and parsers and machine translation systems
\cite{ChurchMercer93}.  In a similar way, annotated
speech corpora have proliferated and have found uses across a rapidly
expanding set of languages, disciplines and technologies
\sburl{www.ldc.upenn.edu/annotation/}.
Over the last 7 years, the Linguistic Data Consortium (LDC)
has published over 150 text and speech databases
\sburl{www.ldc.upenn.edu/Catalog/}.

Typically, such databases are specified at the level of file
formats.  Linguistic content is annotated with a variety of
tags, attributes and values, with a specified syntax and semantics.
Tools are developed for each new format and linguistic domain
on an ad hoc basis.  These systems are
akin to the databases of the 1960s.  There is a physical
representation along with a hand-crafted program offering a
single view on the data.  Recently, the authors have shown how
the three-level architecture and the relational model can be
applied to annotated speech databases
\cite{BirdLiberman99,Cassidy99}.  The goal of this paper is
to illustrate our two approaches and to describe ongoing research
on query algebras.

Before presenting the models we give an example of a collection
of speech annotations.  This illustrates the diversity of
the physical formats and gives an idea of the challenge involved
in providing a general-purpose logical characterisation of the
data.  The Boston University Radio Speech Corpus consists of
7 hours of radio news stories
\sburl{www.ldc.upenn.edu/Catalog/LDC96S36.html}.
The annotations include four
types of information: orthographic transcripts, broad phonetic
transcripts (including main word stress), and two kinds of prosodic
annotation, all time-aligned to the digital audio files. The two kinds
of prosodic annotation implement the system known as ToBI --
Tones and Break Indices
\sburl{www.ling.ohio-state.edu/phonetics/E_ToBI/}.
We have added three further annotations: coreference annotation and
named entity annotation in the style of MUC-7
\sburl{www.muc.saic.com/proceedings/muc_7_toc.html}, and
syntactic structures in the style of the Penn TreeBank \cite{Marcus93}.
Fragments of the physical data are shown in Figure~\ref{fig:bu-speech}.

\begin{figure*}
{\scriptsize\setlength{\tabcolsep}{.5\tabcolsep}
\begin{tabular}{l|l|l}
\begin{minipage}[t]{.325\linewidth}
{\small Coreference Annotation}

{\tiny
\begin{alltt}
<COREF ID="2" MIN="woman">
  This woman</COREF>
receives three hundred dollars
a month under
<COREF ID="5">
  General Relief</COREF>, plus
<COREF ID="16"
       MIN="four hundred dollars">
  four hundred dollars a month in
  <COREF ID="17"
         MIN="benefits" REF="16">
    A.F.D.C. benefits</COREF>
</COREF> for
<COREF ID="9" MIN="son">
  <COREF ID="3" REF="2">
    her</COREF> son
</COREF>, who is
<COREF ID="10" MIN="citizen" REF="9">
  a U.S. citizen</COREF>.
<COREF ID="4" REF="2">
  She</COREF>'s among
<COREF ID="18" MIN="aliens">
  an estimated five hundred illegal
  aliens on
  <COREF ID="6" REF="5">
    General Relief</COREF>
  out of
  <COREF ID="11" MIN="population">
    <COREF ID="13" MIN="state">
      the state</COREF>'s
    total illegal immigrant
    population of
    <COREF ID="12" REF="11">
      one hundred thousand
    </COREF>
  </COREF>
</COREF>.
<COREF ID="7" REF="5">
  General Relief</COREF>
is for needy families and
unemployable adults who
\end{alltt}}
\end{minipage}
&
\begin{minipage}[t]{.25\linewidth}
{\small Named Entity\\ Annotation}

{\tiny
\begin{alltt}
This woman receives
<b_numex TYPE="MONEY">
  three hundred dollars
<e_numex>
a month under General
Relief, plus
<b_numex TYPE="MONEY">
  four hundred dollars
<e_numex>
a month in A.F.D.C.
benefits for her
son, who is a
<b_enamex TYPE="LOCATION">
  U.S.
<e_enamex>
citizen. brth She's among
an estimated five hundred
illegal aliens on General
Relief brth out of the
state's total illegal
immigrant population of
one hundred thousand. brth
General Relief is for
needy families and
unemployable adults brth
who don't qualify for other
 public assistance. brth
<b_enamex TYPE="ORGANIZATION">
  Welfare Department
<e_enamex>
spokeswoman
<b_enamex TYPE="PERSON">
  Michael Reganburg
<e_enamex>
brth says the state will
save about
<b_numex TYPE="MONEY">
  one million dollars
<e_numex>
a year if illegal aliens
are denied General Relief.
\end{alltt}}
\end{minipage}
&
\begin{minipage}[t]{.325\linewidth}
{\small Penn Treebank Annotation}

{\tiny
\begin{alltt}
((S
  (NP-SBJ This woman)
   (VP receives
    (NP
     (NP
      (NP (QP three hundred) dollars)
      (NP-ADV a month)
      (PP under
       (NP General Relief))) , plus
     (NP
      (NP (QP four hundred) dollars
      )
      (NP-ADV a month)
      (PP in
       (NP A.F.D.C. benefits))))
    (PP for
     (NP
      (NP her son) ,
      (SBAR (WHNP-1 who)
       (S (NP-SBJ *T*-1)
        (VP is
         (NP-PRD a U.S. citizen)))))))
  .
))
((S
 (NP-SBJ She)
 (VP 's
  (PP-PRD among
   (NP (NP an estimated
    (QP five hundred) illegal aliens)
   (PP on
    (NP General Relief))
   (PP out of
    (NP
     (NP
      (NP the state 's)
       total illegal immigrant population)
       (PP of
        (NP
         (QP one hundred thousand))))))))
  .
\end{alltt}}
\end{minipage}
\end{tabular}

\vspace*{2ex}\hrule\vspace*{2ex}

\begin{tabular}{l|l|l|l}
\begin{minipage}[t]{.21\linewidth}
{\small Word-Level Annotation}

\begin{alltt}
0.320000 This
0.620000 woman
1.120000 receives
1.370000 three
1.670000 {hundred
2.020000 }dollars
2.060000 a
2.450000 month
2.740000 under
3.280000 General
3.800000 Relief
4.310000 plus
4.520000 four
4.800000 hundred
5.160000 dollars
5.190000 a
5.480000 month
5.610000 in
6.340000 A.F.D.C.
6.870000 benefits
7.060000 for
7.190000 her
7.620000 son
7.830000 who
7.970000 is
8.020000 a
\end{alltt}
\end{minipage}
&
\begin{minipage}[t]{.25\linewidth}
{\small Syllable Annotation}

\begin{alltt}
H#   0    2
H#   2    3
>endsil
DH   5    14   4.182398
IH   19   6   -0.184139
S    25   8   -0.387113
>This
W    33   6   -0.495798
UH+1 39   3   -0.792806
M    42   7    0.042605
>
EN   49   14   0.395379
>woman
R    63   3   -0.996359
IY   66   7   -0.658371
>
S    73   12   0.865892
IY+1 85   13   0.815127
V    98   9    0.815878
Z    107  6   -0.563102
>receives
TH   113  9    0.506469
R    122  5   -0.359288
IY+1 127  11   0.323961
>three
HH   138  3   -0.905714
\end{alltt}
\end{minipage}
&
\begin{minipage}[t]{.33\linewidth}
{\small Tonal Annotation}

\begin{alltt}
0.373684 HiF0
0.493698 H*
0.915000 !H*
1.100000 !H-
1.325000 L+H*
1.389472 HiF0
1.716865 L*
2.178711 !H*
2.434735 L-L%
2.969376 H*
3.552627 HiF0
3.630000 H* ; !HL%, maybe LL% ?
3.770074 H-L%
4.440000 H*
4.478946 HiF0
5.330000 L*
5.445000 L-H%
5.709989 H*
6.300000 H*
6.331575 HiF0
6.740000 L-H%
7.336837 HiF0
7.402120 H*
7.607943 L-L%
8.301393 H*
8.510248 HiF0
10.105260 HiF0
\end{alltt}
\end{minipage}
&
\begin{minipage}[t]{.2\linewidth}
{\small Part-of-speech\\ Annotation}

\begin{alltt}
This DT
woman NN
receives VBZ
three CD
hundred CD
dollars NNS
a DT
month NN
under IN
General NP
Relief, NP
plus CC
four CD
hundred CD
dollars NNS
a DT
month NN
in IN
A.F.D.C. NP
benefits NNS
for IN
her PP\$
son, NN
who WP
is VBZ
\end{alltt}
\end{minipage}
\end{tabular}}

\caption{Multiple Annotations of the Boston University Radio Speech Corpus}\label{fig:bu-speech}
\vspace*{2ex}\hrule
\end{figure*}


Coreference annotation (Figure~\ref{fig:bu-speech}, top left) associates a
unique identifier to each noun phrase and a reference attribute which
links each pronoun to its antecedent.  The set of coreferring
expressions is considered to be an equivalence class.  Named-entity
annotation (top centre) identifies and classifies numerical and name
expressions.  Penn Treebank annotation provides a syntactic parse of
each sentence.  The word-level annotation (bottom left) gives the end
time of each word (a second offset into the associated signal data).
The syllable annotation gives the Arpabet phonetic symbols
(see \sburl{www.ldc.upenn.edu/doc/timit/phoncode.doc}).
The tonal annotation provides time points and intonational units, and the
part of speech annotation (bottom right) specifies the syntactic
category of each word.  This is but a small sample of the bazaar of
data formats.
\arXivhack

\Section{Data Models for Speech Databases}

Two database models for multi-layered speech annotations have been
developed by the authors.  The Emu model (Macquarie) organises the data
primarily in terms of its hierarchical structure, while the annotation
graph model (Penn) foregrounds the temporal structure.  In separate
work we demonstrate the expressive equivalence of the two models
\cite{BirdLiberman99,CassidyHarrington99}.  Here we give a brief
overview of both models. In the remainder of this paper we will
consider mainly the annotation graph data model, while the Emu system
serves as an example of a working speech database system.

\SubSection{The Emu model}

The Emu speech database system \sburl{www.shlrc.mq.edu.au/emu}
\cite{CassidyHarrington96,CassidyHarrington99} provides tools for creation, query
and analysis of data from annotated speech databases.  Emu is
implemented as a core C++ library and a set of extensions to the Tcl
scripting language which provide a set of basic operations on speech
annotations.  Emu provides a flexible annotation model into which a
number of existing label file formats can be read.

The Emu annotation model is based on a set of \emph{levels} which
represent different types of linguistic data such as words, phonemes or
pitch events.  Each level contains a set of \emph{tokens} which have
one or more \emph{labels} and optionally a start and end time relative
to an associated speech signal.  Within a level, tokens are stored as a
partial order representing thier sequence in the annotation: each token
may have zero or more previous and next tokens.  The partial ordering
must respect timing information if it is present in the tokens: that
is, a token cannot follow a token with an later start time.

\begin{figure*}[tbp]
\centerline{\epsfig{file=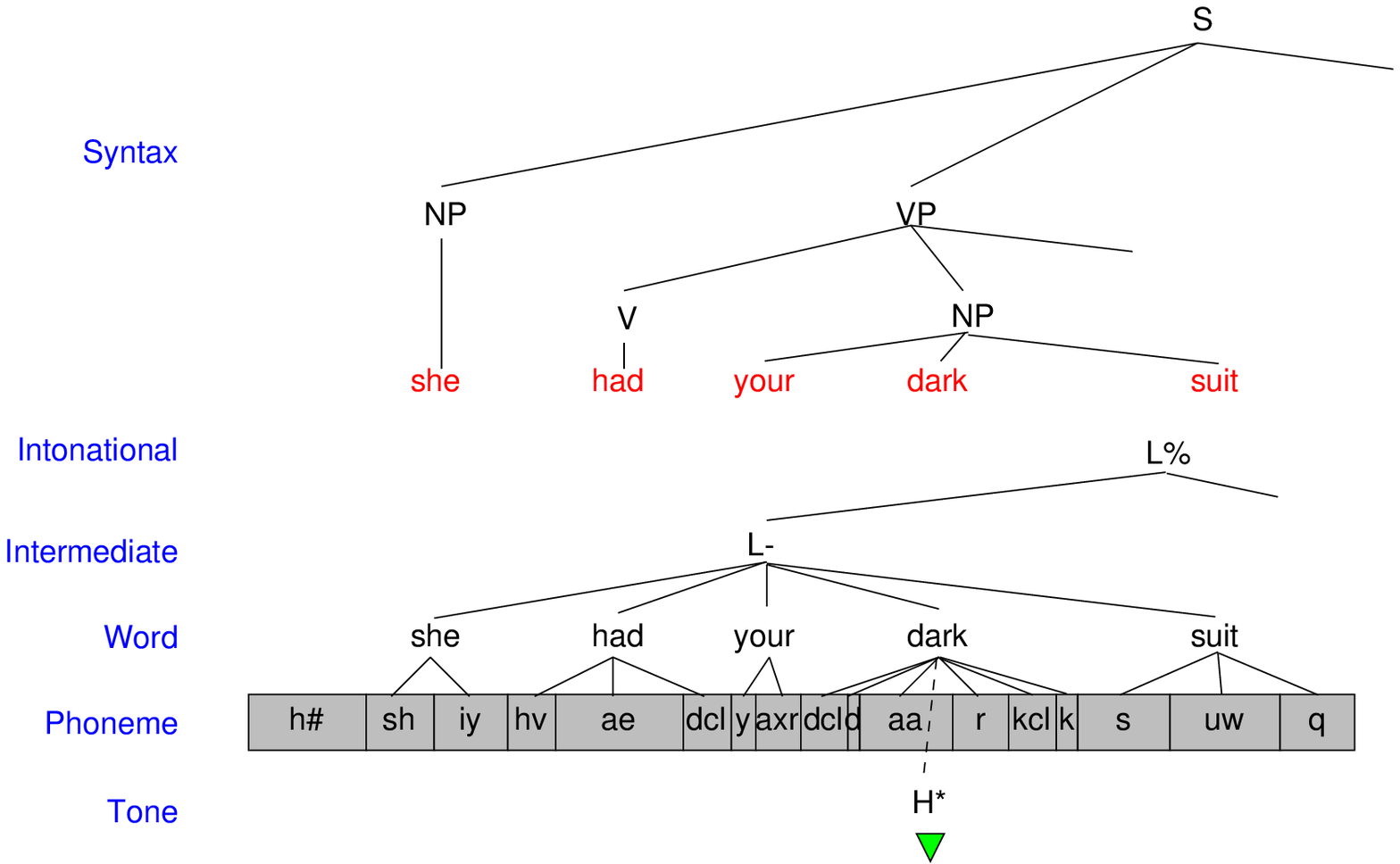,width=0.75\linewidth}}
\caption{An example utterance from the TIMIT database which has been
  augmented with both a syntactic annotation and a ToBI style
  intonational annotation.  The names of the levels are shown on the
  left, the Word level has been duplicated to show the links to both
  the syntactic and intonatational hierarchies. The single Tone event H*
  is associated with the word `dark'. Time information at the phoneme
  level is used to derive times for all higher levels.}
\label{fig:emu-timit}
\vspace*{2ex}\hrule
\end{figure*}

Within and between levels, tokens may be related by either
\emph{domination} or \emph{association} relations.  Domination
relations relate a parent token to an ordered sequence of constituent
child tokens and imply that the start and end times of the parent could
be inferred from those of the children. Association relations have no
in-built semantics and can be used for any application specific
relation, such as that between a word and a tone target which denotes
the point at which word stress is realised
(Figure~\ref{fig:emu-timit}).  Relations may be defined between any
pair of levels which allows Emu to handle intersecting hierarchies such
as that illustrated in Figure~\ref{fig:emu-timit}.

\SubSection{The annotation graph model}

A second general purpose model supporting multiple independent
hierarchical transcriptions of the same signal data is known as the
{\it annotation graph} \cite{BirdLiberman99dtag,BirdLiberman99}.
This model forms the heart of a joint initiative between LDC, NIST
\sburl{www.nist.gov} and MITRE \sburl{www.mitre.org}
to develop an architecture and tools for linguistic
analysis systems (ATLAS), and an NSF-sponsored project between
LDC, the Penn database group, and the CMU Psychology and Informedia
departments, to develop a multimodal database of communicative
interaction called Talkbank \sburl{www.talkbank.org}.

\begin{figure*}[tbp]
\centerline{\epsfig{file=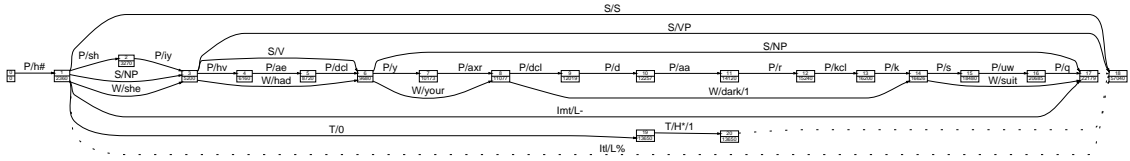,width=\linewidth}}
\caption{TIMIT Graph Structure}\label{fig:ag-timit}
\vspace*{2ex}\hrule
\end{figure*}

Annotation graphs are labelled DAGs with time references on some of the
nodes.  Bird and Liberman have demonstrated that annotation graphs are
sufficiently expressive to encompass the full range of current speech
annotation practice.  A simple example of an annotation graph is shown
in Figure~\ref{fig:ag-timit}, for a corpus known as TIMIT \cite{TIMIT86}.
Annotation graphs (AGs) have the following structure.
Let $L = \bigotimes L_i$ be the label data which occurs on the arcs of
an AG.  The nodes $N$ of an AG reference signal data by virtue of a
function mapping nodes to time offsets $T$.  AGs are now defined as
follows:

\newtheorem{defn}{Definition}
\newtheorem{ex}{Example}

\begin{defn}
An \textbf{annotation graph} $G$ over a label set $L$ and a
timeline $T$ is a 3-tuple
$\left< N, A, \tau \right>$ consisting of a node set $N$,
a collection of arcs $A$ labelled with elements of $L$,
and a time function $\tau$, which satisfies the following conditions:

\begin{enumerate}\setlength{\itemsep}{0pt}

\item $\left< N, A \right>$ is an acyclic digraph
  labeled with elements of $L$, and
  containing no nodes of degree zero;

\item $\tau: N \rightharpoonup T$,
  such that, for any path from node $n_1$ to $n_2$ in $A$,
  if $\tau(n_1)$ and $\tau(n_2)$ are defined, then
  $\tau(n_1) \leq \tau(n_2)$;

\end{enumerate}
\end{defn}

Note that AGs may be disconnected or empty, and that they must
not have orphan nodes.  The AG corresponding to the Emu annotation
structure in Figure~\ref{fig:emu-timit}, for the first five
words of a TIMIT annotation, is given in Figure~\ref{fig:ag-timit}.
The arc types are interpreted as follows:
\expr{S} -- syntax;
\expr{W} -- word;
\expr{P} -- phoneme;
\expr{T} -- tone;
\expr{Imt} -- intermediate phrase;
\expr{Itl} -- intonational phrase.
\arXivhack

\Section{Annotations as Relational Tables}

Annotation data expressed in either the Emu or annotation graph data
models can be trivially recast as a set of relational tables
\cite{Cassidy99}.  For the purposes of this paper it is instructive to
consider the relational form of annotation data in order to explore the
requirements for a query language for these databases.

An annotation graph can be represented as a pair of tables, for the arc
relation and time relations.  The arc relation is a six-tuple
containing an arc id, a source node id, a target node id, and three
labels taken from the sets $L_1, L_2, L_3$ respectively.  The choice of
three label positions is somewhat arbitrary, but it seems to be
both necessary and sufficient for the various annotation structures
considered here.

We let $L_1$ be the set of types of transcript information
(e.g.\ `word', `syllable', `phoneme'), and let
$L_2$ be the substantive transcript element (e.g.\ particular
words, phonetic symbols, and so on).  We let $L_3$ be the names
of equivalence classes, used here to model so-called
`phonological association'.  (This kind of association is
discussed in depth in \cite{Bird95}.)
Let $T$ be the set
of non-negative integers, the sample numbers.
Figure~\ref{fig:graph-table} gives an example for the TIMIT data of
Figures~\ref{fig:emu-timit}, \ref{fig:ag-timit}.

\begin{figure*}[tbp]
{\scriptsize
\begin{minipage}{\textwidth}
\begin{tabular}[t]{c|cccccc}
{\it Arc} &
 $id$ &
 $X$  &
 $Y$  &
 $L_1$ &
 $L_2$ &
 $L_3$ \\
\cline{2-7}

&1 & 0 & 1 & P & h\# & \\
&2 & 1 & 2 & P & sh  & \\
&3 & 2 & 3 & P & iy  & \\
&4 & 3 & 4 & P & hv  & \\
&5 & 4 & 5 & P & ae  & \\
&6 & 5 & 6 & P & dcl & \\
&7 & 6 & 7 & P & y   & \\
&8 & 7 & 8 & P & axr & \\
&9 & 8 & 9 & P & dcl & \\
&10 & 9 & 10 & P & d & \\
&11 & 10 & 11 & P & aa & \\
&12 & 11 & 12 & P & r & \\
&13 & 12 & 13 & P & kcl & \\
&14 & 13 & 14 & P & k & \\
&15 & 14 & 15 & P & s & \\
&16 & 15 & 16 & P & uw & \\
&17 & 16 & 17 & P & q &
\end{tabular}\hfil
\begin{tabular}[t]{c|cccccc}
{\it Arc} &
 $id$ &
 $X$  &
 $Y$  &
 $L_1$ &
 $L_2$ &
 $L_3$ \\
\cline{2-7}

&18 & 1 & 3 & W & she & \\
&19 & 3 & 6 & W & had & \\
&20 & 6 & 8 & W & your & \\
&21 & 8 & 14 & W & dark & 1 \\
&22 & 14 & 17 & W & suit & \\
&23 & 1 & 18 & S & S & \\
&24 &3 & 18 & S & VP & \\
&25 &1 & 3 & S & NP & \\
&26 &3 & 6 & S & V & \\
&27 &6 & 17 & S & NP & \\

&28 &1 & 17 & Imt & L- & \\
&29 &1 & 18 & Itl & L\% & \\

&30 &1 & 19 & T & 0 & \\
&31 &19 & 20 & T & H* & 1
\end{tabular}\hfil
\begin{tabular}[t]{c|cc}
{\it Time} &
 $N$ &
 $T$\\
\cline{2-3}
& 0 & 0    \\
& 1 & 2360 \\
& 2 & 3270 \\
& 3 & 5200 \\
& 4 & 6160 \\
& 5 & 8720 \\
& 6 & 9680 \\
& 7 & 10173\\
& 8 & 11077\\
& 9 & 12019\\
& 10 & 12257\\
& 11 & 14120\\
& 12 & 15240\\
& 13 & 16200\\
& 14 & 16626\\
& 15 & 18480\\
& 16 & 20685\\
& 17 & 22179\\
& 18 & 57040\\
& 19 & 13650\\
& 20 & 13650
\end{tabular}
\end{minipage}    
    \caption{The Arc and Time Relations}
    \label{fig:graph-table}
}
\vspace{2ex}\hrule
\end{figure*}

We form the transitive closure of the (unlabeled) graph relation to
define a structural (graph-wise) precedence relation using a datalog program:

\begin{sv}
s_prec(X,X) :- 
s_prec(X,Y) :- arc(_,X,Y,_,_,_)
s_prec(X,Y) :- s_prec(X,Z),
               arc(_,Z,Y,_,_,_)
\end{sv}

Now we further define a temporal precedence relation, where {\tt leq} is
the $\leq$ relation (minimally defined on the times used by the graph):

\begin{sv}
t_prec0(X,Y) :- time(X,T1),
                time(Y,T2),
                leq(X,Y)

t_prec(X,Y) :-  t_prec0(X,Y)
t_prec(X,Y) :-  t_prec(X,Z),
                t_prec0(Z,Y)
\end{sv}
\arXivhack

\Section{Exploring Annotated Linguistic Databases}

\SubSection{General architecture}

In our experience with the analysis of linguistic databases, we have
found a recurrent pattern of use having three components
which we will call query, report generation, and analysis.

The query system proper can be viewed as a function from annotation
graphs to sets of subgraphs, i.e. those meeting some (perhaps complex)
condition.
The report generation phase is able to access these query
results, but also the signals underlying the annotations.  For
example, the report generation phase can calculate such things as
`mean F$_2$ in signal S during time interval $(t_1,t_2)$.'
Each hit constitutes an `observation' in the statistician's sense,
and we extract a vector of specified values for each observation, to
be passed along to the analysis system.
The analysis phase is then some general-purpose data
crunching system such as Splus or Matlab.

This architecture saves us from having to incorporate all possible
calculations over annotated signals into the query language.
The report generation phase can perform such calculations, as well
as compute properties of the annotation data itself.
This seems to simplify the query system a good deal;
now things like `count the number of syllables to the end of the
current phrase' (which we do need to be able to do) are tasks for the
report generator, not the query system proper.

In general, the result of a query is a set of sub-graphs, each of which 
forms one matching instance.  If we use the relational model proposed
above, these would be returned as a result table having the 
same structure as the arc relation of Figure~\ref{fig:graph-table},
but containing just the tuples which took part in each matching instance. 
We are then faced with the problem of how to differentiate the matching 
instances, for example, if we wished to collect together the word
labels for the query `find all words dominated by noun phrases' we need 
some way of treating each sub-graph separately.  Hence, we would prefer 
the result to be a set of tables rather than a single table containing
all matching tuples. 

In a sense, then, the only role of the query is to define an iterator
for the report generator over a set of sub-graphs of the overall
annotation graph.

\subsubsection*{The Emu query language}

The Emu query language uses simple conditions on token labels which
match only tokens at a specified level, for example:
\texttt{Phonetic=A|I|O|U|E|V}.  These conditions can be combined by
sequence, domination or association operators to constrain the
relational structure of the tokens of interest.  Examples of each are:

  Find a sequence of \texttt{vowel} followed by \texttt{stop} at the
  phoneme level:\\
      \texttt{[Phoneme=vowel \queryseq Phoneme=stop]}
 
  Find Words not labelled \texttt{x} dominating \texttt{vowel}
  phonemes:\\
      \texttt{[Word!=x \querydom Phoneme=vowel]}

  Find words associated with \texttt{H*} tones:\\
      \texttt{[Word!=x \queryassoc Tone=H*]}     
     
The \texttt{Word!=x} query is intended to match any word in lieu of a
query language construct which allows matching any label string.  

Each query matches either a token or, in the case of the sequence
query, a sequence of tokens.  The result of a domination or association
query is the result of the left hand side of the bracketed term; this
can be changed by marking the right hand side term with a hash (\texttt{\#}).
Compound queries can be arbitrarily nested to specify complex
constraints on tokens. As an example the following query finds
sequences of stop and vowels dominated by strong syllables where the
vowel is associated with an \texttt{H*} tone target, the result is a
list of the vowel labels with associated start and end times.
\begin{center}
\begin{sv}
 [Syllable=S ^ 
     [Phoneme=stop -> 
         [Phoneme=vowel => Tone=H*]]]
\end{sv}
\end{center}

The result of an Emu query is a table with one entry per matching
token:
\begin{sv}
database:timit
query:Phoneme!=x
type:segment
#
h#      0       147.5   fjsp0:sa1
sh      147.5   232.5   fjsp0:sa1
iy      232.5   325     fjsp0:sa1
hv      325     385     fjsp0:sa1
...
\end{sv}
This table is used to extract any of the associated time-series data
associated with the database, an operation usually carried out from an
analysis environment such as Splus or XlispStat.  Emu provides
libraries of analysis functions for these environments which
facilitate, for example, mapping signal processing operations over each
token in a query result or overlaying plots of the time series data for
each token.

Although this query system has proved useful and useable in the
environment of acoustic phonetics research, it is now evident that
there are a number of shortcomings which prevent it's wider use. The
query syntax is unable to express some queries, such as those involving 
disjunction or optional elements, and the query result is only really
useful for data extraction.  It is for these reasons that we are now
looking more formally at the requirements for a query language for
annotation data. 

\SubSection{A query language on annotation graphs}

A high-level query language for annotation graphs, founded on
an interval-based tense logic, is currently being developed and
will be reported in a later version of this paper.

Here we describe a variety of useful queries on annotation
graphs and formulate them as datalog programs.  As we shall see,
it turns out that datalog is insufficiently expressive for the
range of queries we have in mind.  Finding a more expressive
yet tractable query language is the focus of ongoing research.

A number of simple operations, extending our two relations
\predicate{arc/6} and \predicate{time/2},
will be necessary for succinct queries.
The first and most obvious is for hierarchy.  Observe in
Figure~\ref{fig:ag-timit} that there is a notion of structural
inclusion defined by the arcs.  We formulate this as follows:

\begin{sv}
s_incl(I,J) :- 
   arc(I,W,Z,_,_,_), 
   arc(J,X,Y,_,_,_), 
   s_prec(W,X), s_prec(Y,Z)
\end{sv}

Now, since \predicate{s\_prec} is reflexive, so is \predicate{s\_incl}.
Observe that nodes 3 and 6 in Figure~\ref{fig:ag-timit} are connected
by both an \smtt{S/V} arc and a \smtt{W/had} arc.  The syntactic
verb arc \smtt{S/V} should dominate the word arc \smtt{W/had}, but not
vice versa.  Therefore we need to have a hierarchy defined over the
types.  We achieve this with a (domain-specific) ordering on the
type names:

\begin{sv}
type_hierarchy(word,syl)
type_hierarchy(syl,seg)
\end{sv}

\noindent
Now dominance is expressed by the predicate:

\begin{sv}
dom(I,J) :- 
   arc(I,_,_,L1,_,_), 
   arc(J,_,_,L2,_,_), 
   type_hierarchy(L1,L2), 
   s_incl(I,J)
\end{sv}

In some cases it is necessary to have an intransitive dominance
relation that is sensitive to phrase structure rules.  For simplicity
of presentation, we assume binary branching structures.  The first
of the rules below states that a sentence arc \smtt{s} will
immediately and exhaustively dominate an \smtt{np} arc followed
by a \smtt{vp} arc.

\begin{sv}
ps_rule(s,np,vp)
ps_rule(np,det,n)
ps_rule(vp,v,np)
\end{sv}

\noindent
Now we define immediate dominance over the syntax arcs \smtt{syn} as
follows:

\begin{sv}
i_dom(I,J) :- 
   arc(I,X,Z,syn,P,_), 
   ps_rule(P,C1,C2), 
   arc(J,X,Y,syn,C1,_), 
   arc(_,Y,Z,syn,C2,_)

i_dom(I,J) :- 
   arc(I,X,Z,syn,P,_), 
   ps_rule(P,C1,C2), 
   arc(_,X,Y,syn,C1,_), 
   arc(J,Y,Z,syn,C2,_)
\end{sv}

Another widely used relation between arcs is association.  In the
instance of the AG model in Figure~\ref{fig:graph-table}, association
amounts to sharing the value of $L_3$, as we saw in the tuples
for \smtt{dark} and \smtt{H*} in Figure~\ref{fig:graph-table}.
The \predicate{assoc}
predicate simply does a join on the third label field:

\begin{sv}
assoc(I,J) :- 
   arc(I,_,_,_,_,A), 
   arc(J,_,_,_,_,A)
\end{sv}

Finally, it is convenient to have a kleene star relation.
Unfortunately in datalog we are unable to collect up the arbitrary length
sequence it matches.  Here we have it returning the two nodes which
bound the sequence, which is often enough to uniquely identify the
sequence in practice.

\begin{sv}
node(N) :- arc(_,N,_,_,_,_)
node(N) :- arc(_,_,N,_,_,_)

kleene1(X,X,_) :- node(X)
kleene1(X,Y,L) :- 
   arc(_,X,Z,L,_,_),
   kleene1(Z,Y,L)

kleene2(X,X,_) :- node(X)
kleene2(X,Y,L) :- 
   arc(_,X,Z,_,L,_), 
   kleene2(Z,Y,L)

kleene3(X,X,_) :- node(X)
kleene3(X,Y,L) :- 
   arc(_,X,Z,_,_,L), 
   kleene3(Z,Y,L)
\end{sv}

With this simple machinery we can start defining some annotation
queries.

%
%

\noindent
Find a sequence of vowel followed by stop at the phoneme level
(assumes suitably defined {\tt vowel} and {\tt stop} unary relations):
\begin{sv}
vowel_stop(I,J) :- 
   arc(I,_,Y,phoneme,V,_), 
   arc(J,Y,_,phoneme,S,_), 
   vowel(V), stop(S)
\end{sv}

\noindent
If we do not want both the vowel and the stop, but just the vowel,
we could write:
\begin{sv}
vowel_stop(I) :- 
   arc(I,_,Y,phoneme,V,_), 
   arc(_,Y,_,phoneme,S,_), 
   vowel(V), stop(S)
\end{sv}

\noindent
Find words dominating vowel phonemes:
\begin{sv}
strongWrdDomVowels(I) :-
    arc(I,_,_,word,s,_),
    arc(J,_,_,phoneme,V,_),
    vowel(V),
    dom(I,J)
\end{sv}

\noindent
Find words associated with H* tones:
\begin{sv}
sylHtone(I) :-
    arc(I,_,_,word,_,A),
    arc(_,_,_,tone,h*,A)
\end{sv}

\noindent
Find stop-vowel sequences dominated by words in noun phrases where the word
is associated with an H* tone target.
\begin{sv}
stop_vowel_seq(I,J) :-
    arc(I,_,Y,phoneme,S,_), stop(S),
    arc(J,Y,_,phoneme,V,_), vowel(V),
    arc(W,_,_,word,_,_),
    arc(N,_,_,syn,np,_),
    dom(N,W), dom(W,I), dom(W,J),
    arc(T,_,tone,h*,_), assoc(W,T)
\end{sv}

\noindent
Find the intermediate phrase containing the main verb of a sentence:
\begin{sv}
imt_phrase(P) :-
    arc(K, _, _, syn, s, _),
    arc(J, _, _, syn, vp, _),
    arc(I, _, _, syn, v, _),
    i_dom(K,J), i_dom(J,I),
    dom(P, I),
    arc(P, _, _, imt, _, _)
\end{sv}

\noindent
Return the set of syllables between an H* and an L\% tone (inclusive).
\begin{sv}
syls(K) :-
    arc(_, _, N, tone, h*, A1),
    arc(_, N, _, tone, l%, A2),
    arc(I, _, N1, syl, _, A1),
    arc(J, N2, _, syl, _, A2),
    kleene1(N1, N2, syl),
    arc(K, N2, N3, syl,_,_),
    kleene1(N3, N4, syl)
\end{sv}

The above query shows how the datalog model breaks down.  We would
like it to return sets of sets of syllable arcs.  Instead it returns
a flat set structure.
In many cases we will know that some arc participating in
the query expression can be used to recover the nested structure.  For
example, if the head of the above clause was changed from
\predicate{syls(K)} to \predicate{syls(I,K)}, then \predicate{I} will
aggregate \predicate{K} in just the right way.
\arXivhack

\Section{Applying XML Query Languages to Annotations}


It is worth briefly considering the suitability of existing XML
query languages such as XML-QL \cite{xml-ql} and XQL \cite{xql}
for the domain of annotated speech.  At first glance the problems
we face querying annotated speech data are similar to those present
with XML queries in that both present a hierarchical data
model.  A number of formulations of annotation data as XML are
possible, indeed some projects make use of XML/SGML based formats
entirely (e.g.\ MATE \sburl{mate.nis.sdu.dklpq},
LACITO \sburl{lacito.vjf.cnrs.fr/ARCHIVAG/ENGLISH.htm}).
%
%
XML can represent trees using properly nested tags, in the
obvious way.  In order to represent multiple independent
hierarchies built on top of the same material one must construct trees
using IDREF pointers.  This idea was proposed by the
Text Encoding Initiative \cite{TEI-P3} and recently
adopted by the MATE project.  We believe this approach is
vastly more expressive than necessary for representing speech
annotations, and we prefer a more constrained approach having
desirable computational properties with respect to creation,
validation and query.

The XQL proposal \cite{xql} describes a query language which is
intended to select elements from
within XML documents according to various criteria; for example, the
query \texttt{text/author} returns all author elements that are
children of text elements.  The XQL data model ignores the order of
elements within a parent element and has no obvious way to query for
sequences of tokens. 

The XML-QL proposal \cite{xml-ql} provides for a data model
where the order of elements is respected.  A query for a word-internal
vowel-stop sequence could be expressed as follows (assuming
suitably tagged annotation data for TIMIT):

\begin{sv}
<word>
  <phoneme label=&vowel;/>
  <phoneme label=&stop;/>
</word>
\end{sv}

\noindent
The result of this query would have the following form:

\begin{sv}
<word label=had>
  <phoneme label=ae/>
  <phoneme label=dcl/>
</word>
<word label=dark>
  <phoneme label=ar/>
  <phoneme label=k/>
</word>
\end{sv}

Queries which refer to two independent
hierarchies, such as syntactic and intonational phrase
structure, need to use joins.
For example, to find words that are simultaneously
at the end of both clauses and intermediate phrases,
we could have the following query:
\begin{sv}
<intermediate>
  <word id=\$i></>[end()]
</intermediate>

<clause>
  <word id=\$i></>[end()]
</clause>
\end{sv}

We assume the existence of some mechanism to pick out the last child
element.
The ID attribute ensures that the words are the same in each
part of the join.

Perhaps either of these approaches could be made to work for a useful range
of query needs.  However they do not appear to be sufficiently
general.  For example, it is often useful to have query expressions
involving kleene star: `select all pairs of consonants, ignoring
any intervening vowels' (CV*C).  Such queries may ignore hierarchical
structure, finding sequences across (say) word boundaries.
Using regular expressions over paths, XML-QL could provide access to
strings of terminal symbols ignoring intervening levels of hierarchy.
Yet it does not provide regular-expression matching over those
sequences.  Alternatively, sequences at each level of a hierarchy
could be chained together using IDREF pointers, but it is unclear
how we would manage closures over such pointer structures.

\arXivhack

\Section{Conclusions}

Annotated speech corpora are an essential component of speech
research, and the variety of formats in which they are distributed has
become a barrier to their wider adoption.  To address this issue,
we have developed two data
models for speech annotations which seem to be sufficiently expressive
to encompass the full range of practice in this area.  We have shown
how the models can be stored in a simple relational format, and how
many useful queries in this domain are first-order.  However, existing
query languages lack sufficient expressive power for the full range of
queries we would like to be able to express, and we hope stimulate new
research into the design of general purpose query languages for
databases of annotated speech recordings.
\arXivhack

\section*{Acknowledgements}

We are grateful to Peter Buneman, Mark Liberman
and Gary Simons for helpful discussions concerning
the research reported here.

\raggedright\small
\bibliographystyle{latex8}

\end{document}